\pdfoutput=1

\documentclass[11pt,table]{article}

\usepackage{acl}

\usepackage{times}
\usepackage{latexsym}

\usepackage[T1]{fontenc}

\usepackage[utf8]{inputenc}

\usepackage{microtype}

\usepackage{amsmath}
\usepackage{amssymb}
\usepackage{graphicx}
\usepackage{booktabs}
\usepackage{xcolor}
\usepackage{bm}
\usepackage{color, colortbl}
\usepackage{xurl}
\definecolor{Gray}{gray}{0.9}

\DeclareMathAlphabet\mathbfcal{OMS}{cmsy}{b}{n}

\title{Long-term Control for Dialogue Generation: Methods and Evaluation}
\date{}

\author{Ramya Ramakrishnan \\
  ASAPP \\
  \texttt{\normalsize rramakrishnan@asapp.com} \\\And
  Hashan Buddhika Narangodage \\
  ASAPP \\
  \texttt{\normalsize hnarangodage@asapp.com} \\\AND
  Mauro Schilman \\
  ASAPP \\
  \texttt{\normalsize mschilman@asapp.com} \\\And
  Kilian Q. Weinberger \\
  ASAPP, Cornell \\
  \texttt{\normalsize kweinberger@asapp.com } \\\And
  Ryan McDonald \\
  ASAPP \\
  \texttt{\normalsize rmcdonald@asapp.com} \\
  }
 
\begin{document}
\maketitle
\begin{abstract}

Current approaches for controlling dialogue response generation are primarily focused on high-level attributes like style, sentiment, or topic. In this work, we focus on \textit{constrained long-term} dialogue generation, which involves more fine-grained control and requires a given set of control words to appear in generated responses. This setting requires a model to not only consider the generation of these control words in the immediate context, but also produce utterances that will encourage the generation of the words at some time in the (possibly distant) future. We define the problem of constrained long-term control for dialogue generation, identify gaps in current methods for evaluation, and propose new metrics that better measure long-term control. We also propose a retrieval-augmented method that improves performance of long-term controlled generation via logit modification techniques. We show through experiments on three task-oriented dialogue datasets that our metrics better assess dialogue control relative to current alternatives and that our method outperforms state-of-the-art constrained generation baselines. \footnote{Our code is available at \url{https://github.com/asappresearch/constrained-dialogue-generation}}

\end{abstract}

\section{Introduction}

Despite recent advances in dialogue systems \cite{serban2016building,ham2020end}, \textit{controlling} dialogue generation remains a significant challenge. Response generation in dialogue can be controlled towards different topics and styles \cite{madotto2020plug} or towards a set of hard constraints (i.e., lexical control words need to appear in the generated text) \cite{sha2020gradient}. We focus on the hard constraint setting, also known as \textit{constrained} generation, as this provides a more fine-grained method of controlling dialogues.

\begin{figure}[t]
    \centering
    \includegraphics[width=0.49\textwidth]{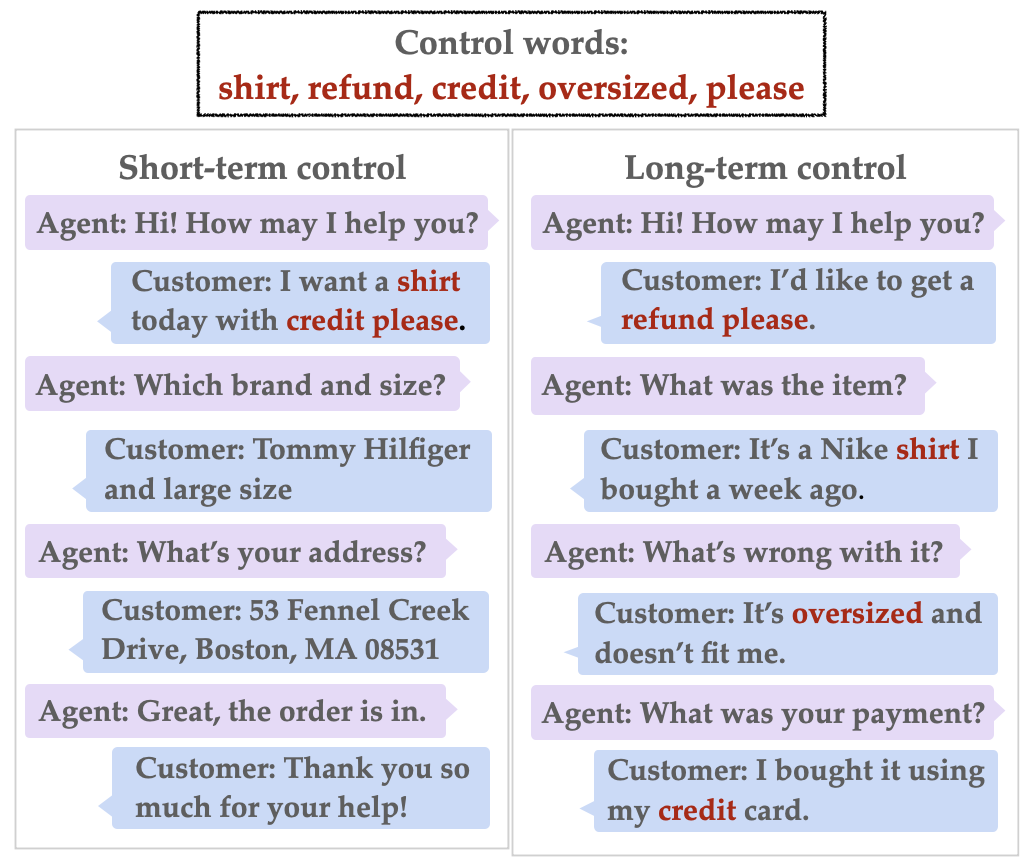}
    \caption{Examples of short vs. long-term control for dialogue generation. \textbf{(Left)} In short-term control, many control words are generated initially, but the conversation is led away from the desired future. \textbf{(Right)} In long-term control, responses are generated with the future in mind with words generated at natural points in the conversation.}
    \label{fig:intro_fig}
\end{figure}

For example, consider a customer service use case (Figure~\ref{fig:intro_fig}), in which an agent speaks to a customer about an issue. The goal is to generate a given set of control words in the responses of one of the speakers (agent or customer). Naive constrained generation approaches \cite{pascual2020directed,miao2019cgmh} use methods like beam search and stochastic search to force the generation of these control words for short-term control, where control words need to appear in a single utterance or phrase. Because they do not consider the future, these approaches may generate the words all at once in a single response or not generate them at natural places in the conversation (Figure~\ref{fig:intro_fig}, left).

The above example highlights the challenges of applying existing constrained generation methods to long-term dialogue generation. First, since another speaker is involved in the dialogue, the model does not have full control of the generated text. Instead, the model can only control the dialogue indirectly. Second, dialogues can be long and thus, controlling utterances several time steps into the future is non-trivial. In this work, we propose the problem of long-term dialogue control, where the goal is to generate a set of control words over many utterances in a dialogue, which requires appropriately timing the generation of control words (Figure~\ref{fig:intro_fig}, right). To the best of our knowledge, we are the first work to constrain \textit{long-term dialogue generation} through lexical control words. 

We begin by highlighting challenges with evaluation for this problem. Successful long-term control of dialogue can be difficult to measure. We describe current evaluation metrics for constrained text generation and show that these metrics can be gamed by generating all or many control words early in the conversation. To resolve this and measure how natural the control is, we propose a new set of metrics: long-term success rate, which measures the percentage of control words in simulated roll-outs of the conversation, and precision, recall, and F1-score, which compare control words in generated responses to those in reference responses from a historical dataset. The second set of metrics specifically help to capture whether the control words are generated at the right time.

Next, we propose a novel method to explicitly address long-term control. Prior methods are unable to handle this task as the number of possible future sequences is exponential. To alleviate this issue, we retrieve similar conversations from training and condition on them during generation. We first identify similar neighbors using a $k$NN-based approach and then guide the language model towards generating similar responses, inspired by plug-and-play methods \cite{madotto2021few,dathathri2019plug,pascual2020directed}. The motivation for this is that retrieved conversations guide the model to generate the control words at more natural points in the conversation.

We conduct experiments on multiple task-oriented dialogue datasets and show that our method outperforms several constrained text generation baselines on automated evaluation metrics as well as human evaluation. Specifically, we are able to generate 30-40\% more control words on long-term success rate compared with baselines, while preserving fluency (scores of $\geq$ 4.3 out of 5), as measured by human evaluation.

\section{Related work}
\paragraph{Controllable text generation.}
Prior work has developed many methods for controllable text generation. These approaches can be categorized into three general areas. The first is altering decoding strategies \cite{NEURIPS2019_d76d8dee,deng2020residual}, in which the sampling distribution can be modified \cite{ghazvininejad-etal-2017-hafez,baheti2018generating} or hidden states in the models can be changed \cite{gu2017trainable}. The second area involves including prompts to guide text generation \cite{ribeiro2018semantically,jiang2020can,li2021prefix}, for example through universal trigger tokens \cite{wallace2019universal,shin-etal-2020-autoprompt}. Finally, fine-tuning can be used to guide language model outputs through the use of a latent variable \cite{fan2018hierarchical,peng2018towards} or through CTRL codes \cite{keskar2019ctrl}. Our work differs from the broad area of controllable language generation in that 1) we require more fine-grained generation through lexical control words and 2) we focus on dialogue settings where another speaker can also change the course of the conversation.
\paragraph{Constrained text generation.}
The key difference between constrained text generation and controllable text generation is the focus on hard rather than soft constraints. Typically, there are two general methods for constrained generation: beam search \cite{hokamp2017lexically,post2018fast,pascual2020directed} and stochastic search \cite{miao2019cgmh,sha2020gradient}. Directed Beam Search (DBS) \cite{pascual2020directed}, modifies language model logits to encourage generation of a specified set of ``guide words", or control words. A method based on stochastic search \cite{miao2019cgmh} uses Metropolis-Hastings with the constraint of keyword inclusion. These approaches do not apply to the dialogue setting where these constraints need to hold for many utterances into the future.
\paragraph{Dialogue response generation.}
While many works develop methods for unconstrained response generation \cite{budzianowski2019hello,peng2020soloist,cao2020pretrained,hosseini2020simple,yavuz2019deepcopy}, there is a subset of work more related to our problem focused on \textit{controlling} response generation. In one work, transformer models are fine-tuned for dialogue through modifications of the inputs, for example by adding information about the user's persona \cite{wolf2019transfertransfo}. The work of \newcite{lippe2020diversifying} generates utterances by paraphrasing templated responses. Several works control generation through exemplar-guided methods \cite{cai2020exemplar,gupta2020controlling}, which is a different setting from ours since we want to guide generation based on a set of control words rather than through a prototype. One work \cite{xu2019neural} controls response generation through meta-words that include desired attributes of the response (e.g., response length and specificity). Another work controls response generation through control words by adding inductive biases into training to guide generation \cite{wu2020controllable}. However, this work only controls generation for a single response, rather than controlling several utterances into the future. The closest work to ours is work by \cite{tang-etal-2019-target}, which proposes a similar problem of long-term control towards a target subject. While the setup is similar, we learn to constrain dialogue responses given a set of control words rather than a target attribute, which also results in a different approach.
\paragraph{Retrieval-augmented generation.}
Another related area is retrieval-augmented language generation, which inspires our approach of using retrieval to control dialogue generation. REALM \cite{guu2020realm} uses a latent knowledge retriever to identify relevant documents and backpropagates through this retrieval step. In another work \cite{fan2020augmenting}, relevant information is retrieved from an external knowledge base to guide dialogue generation. Several works by Khandelwal et al leverage nearest neighbor approaches to improve performance with no additional training \cite{khandelwal2019generalization, khandelwal2020nearest}. While these works condition on retrieval for uncontrolled generation, we leverage ideas from this space specifically for control in dialogue.

\section{Problem definition}
We first define the problem of long-term constrained dialogue generation. A conversation $\mathcal{X} = \{s_1, u_1, s_2, u_2, ..., s_T, u_T\}$ is defined as a list of utterances generated by two speakers: the system $s$ that we are trying to control and the user $u$, which we don't have explicit control over. $T$ denotes the total number of turns in the conversation. Given the current dialogue context of a conversation $x = \{s_1, u_1, ..., s_t, u_t\}$ up until timestep $t$ and a set of control words $\mathcal{W} = \{w_1, w_2, ..., w_M\}$, our goal is to generate the remaining responses of the conversation $\mathcal{R}_{t+1:T} = \{s_{t+1}, ..., s_{T}\}$ such that the control words $\mathcal{W}$ appear in the future generated responses. We consider a scenario in which someone provides a set of control words to be included in the conversation without assumptions on their order. This means methods need to handle control words given in any order.

We additionally assume access to a historical dataset of conversations $\mathcal{D} = \{x^{(i)}\}$, $i \in [1,...,N]$ and a fine-tuned language model $M$ on this dataset. We can leverage these inputs in order to control future responses $\mathcal{R}_{t+1:T}$. We focus on the plug-and-play setting \cite{pascual2020directed}, in which approaches simply guide the given language model $M$ towards generating the control words without any additional re-training.

\section{Proposed metrics for evaluation}
\label{sec:metrics}

Directly evaluating the generated responses in terms of prior evaluation methods can lead to misleading results. Previous works on constrained text generation \cite{pascual2020directed} have used metrics like perplexity to measure fluency and success rate to measure the percentage of control words generated. However, these metrics are more relevant for short-term generation, as they can be gamed in settings where the control words would be naturally distributed across the full conversation. As shown in the left-hand side of Figure \ref{fig:intro_fig}, when several words are forced into the first response, the conversation may move away from the desired future and control word generation could be inappropriately timed. To better evaluate how well the model generates the right words at the right time, we propose the following new metrics.

\begin{figure*}[t]
    \centering
    \includegraphics[width=0.8\textwidth]{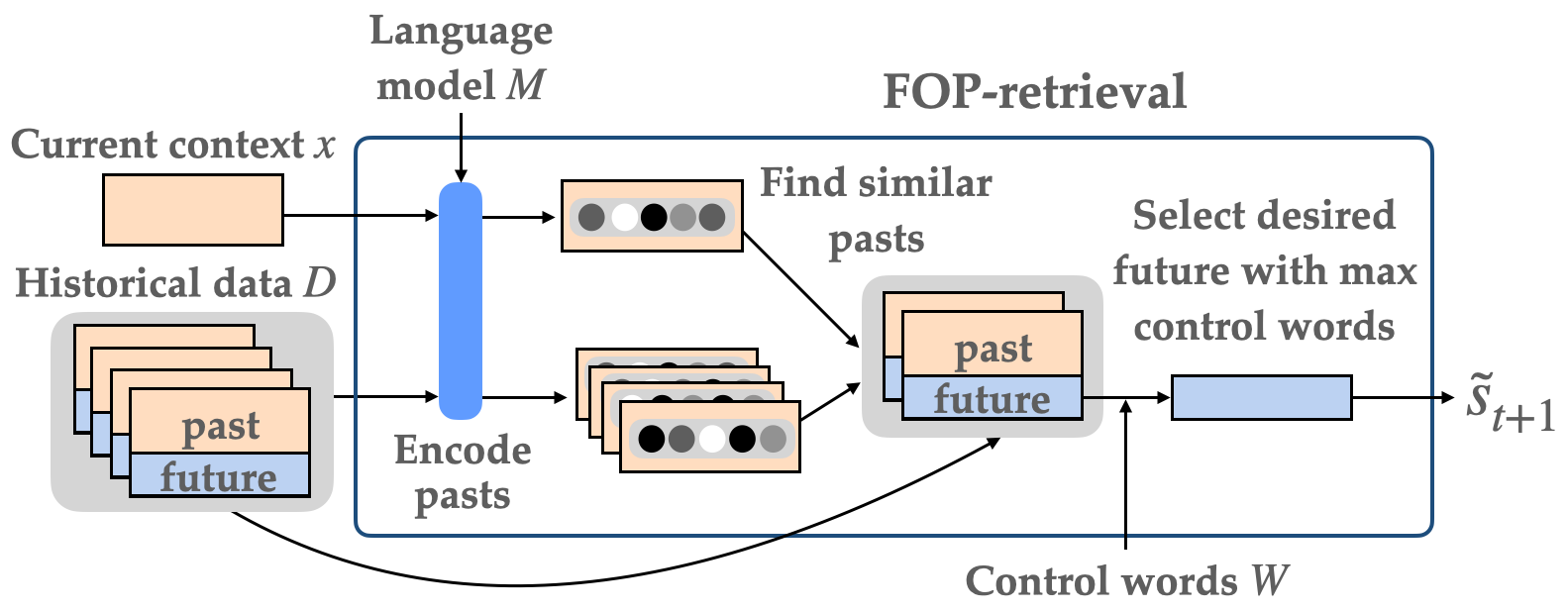}
    \caption{Visualization of FOP-retrieval. First, each conversation in the historical dataset $\mathcal{D}$ is split into many past-future conversation pairs. The current context $x$ and the pasts are encoded using language model $M$. We use $k$NN search to identify pasts similar to context $x$ and then select a desired future with the highest number of control words. The output is the first response in the selected future $\tilde{s}_{t+1}$.}
    \label{fig:fop_retrieval}
\end{figure*}

The first metric we propose is long-term success rate, which involves simulating conversations with a user language model and computing the percentage of generated control words in the system responses of these simulated roll-outs. Prior work \cite{ghandeharioun2019approximating} has used self-play for evaluation, but they do not propose roll-outs as a way to measure dialogue control.
\paragraph{Long-term success rate:} Our modified success rate metric is computed as the fraction of control words generated in a full simulated roll-out of the conversation. We compute this as: $s = \frac{n_w}{|\mathcal{W}|}$, where $n_w$ is the number of control words that appear in all of the future system responses $\mathcal{R}_{t+1:T}$.

One limitation of long-term success rate is that it doesn't measure the timing of control words in the conversation. So next, we want to evaluate whether the methods generate control words at appropriate points in the conversation. To measure this, we propose computing precision, recall, and F1-score for control words. This particular evaluation is not done in simulation. Instead, we consider each true system response in the evaluation dataset in isolation and generate a response for each, given the conversation history up until that point. We compute the number of generated control words that are correctly predicted, when compared with the control words in the ground truth response \emph{in the same time step}.

For example, on the right side of Figure 1, when generating the second customer response (given the true conversation history up until then), we would count a ``correct" prediction for P/R/F1 as a response that includes the word ``shirt" (in any position in the response), as it is a control word that appears in the ground truth response in that time step. It is true that control words can also appear later in the conversation, but this setting is already evaluated by long-term success rate in simulated rollouts. After counting the number of correctly predicted control words for each response individually, we aggregate across all responses.
\paragraph{Precision:} Precision is calculated at the corpus-level as the number of correctly predicted control words over the total number of predicted control words ($p = \frac{|\text{correct}|}{|\text{predicted}|}$).
\paragraph{Recall:} Recall is similarly computed at the corpus-level as the number of correctly predicted control words over the total number of actual control words ($r = \frac{|\text{correct}|}{|\text{actual}|}$).
\paragraph{F1-score:} Finally, F1-score combines precision and recall into one metric ($f1 = \frac{(2*p*r)}{(p+r)}$).

These metrics penalize models that condense all control words into one response. Instead, we want the models to naturally generate control words when they are relevant. These metrics evaluate whether control words are generated at the appropriate position in a conversation. To introduce some flexibility, an extension could be to compute a soft version of precision, recall, and F1-score that scores utterances based on whether control words appear within N utterances of the ground truth position. 

Finally, we use human evaluation to evaluate how realistic and relevant the generated responses are. Specifically, we evaluate each conversation on fluency, consistency of control word generation, relevance, coherence, and diversity.

\begin{figure*}[t]
    \centering
    \includegraphics[width=1.01\textwidth]{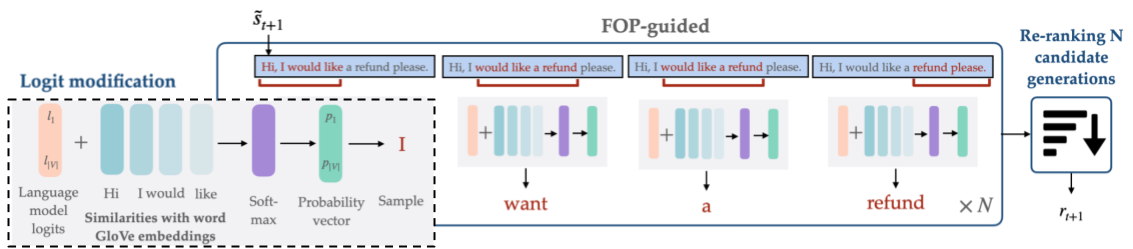}
    \caption{Visualization of FOP-guided. Language model logits are first modified using a window-based approach. All words (and similar words based on GloVe vector similarity) within the window are upweighted with a weight decay. Once any word in the window is generated, the window shifts until the full response is generated. After $N$ generations, a re-ranking step selects the response with the highest number of control words and lowest loss.}
    \label{fig:fop_guided}
\end{figure*}

\section{Retrieval-based Control}

We now present our proposed approach for constrained dialogue generation. Inspired by work in retrieval-augmented generation \cite{guu2020realm,fan2020augmenting}, we retrieve similar pasts based on the current context $x$ and use their futures to control dialogue response generation. The key insight here is that by looking at how people have used these control words in similar conversations in the past, we can bias the models towards more natural dialogues. In other words, we use futures of the past conversations to guide the current response generation. To better motivate the use of retrieval in our problem, consider the example conversation in Figure 1. The agent asks which item the customer wants to return, and there are many possible answers (e.g., ``I want my pant refunded.", ``I want to return gloves I bought yesterday."). Keyword-based retrieval will surface a response about shirts, a control word, which encourages the model to generate a natural response with that word: ``It's a Nike shirt I bought a week ago."

We present two variants of our retrieval-inspired Futures of the Past (FOP) approach: 1) FOP-retrieval: we retrieve the desired future from historical data and simply use the retrieved utterance as the generated response and 2) FOP-guided: we use the utterance from FOP-retrieval as a reference sentence to guide the model towards similar responses. 

The simple variant of our approach, FOP-retrieval, is shown in Figure \ref{fig:fop_retrieval}. It focuses on identifying what the model should say now that will lead to the control words in the future. The reason we need to determine what to say now is that control words in our problem are distributed across a long dialogue conversation. One possible approach to generate the current response is to run many roll-outs of the conversation and select the response that leads to the highest number of control words. However, this brute force approach is computationally expensive and will not be effective for rich, diverse conversations. Instead, we leverage historical conversation data to identify the most relevant futures given the current context and control words. The retrieved futures can guide the model towards what to say now that will lead to the desired future. The guided variant, shown in Figure \ref{fig:fop_guided}, involves guiding the language model towards generating a response similar to the retrieved utterance. 

Our proposed approaches address some of the challenges of long-term control for dialogue generation. First, another speaker can change the course of the conversation, which is why we retrieve a new set of similar past contexts at each time step to re-align with the current context. Second, to control responses many steps into the future, we retrieve historical conversations with the desired future (high percentage of control words) and gently nudge the conversation in that direction, thus controlling not only the current utterance but also the future of the conversation.

\subsection{Retrieval Futures of the Past (FOP-retrieval)}

For the retrieval component, the goal is to select futures that have relevant past contexts as well as desired futures based on the control words. To do this, we employ a multi-step approach. First, we split each conversation $x^{(i)}$ in the historical dataset $\mathcal{D}$ into a set of past-future conversation pairs $x^{(i)} = \{(p,f)^{(i,j)}\}$. We encode the current context $M(x)$ and each past conversation $M(p^{(i,j)})$ using the language model $M$. Then, we use $k$NN search based on FAISS, a library for fast nearest neighbor retrieval \cite{johnson2019billion}, to identify $k$ similar pasts from the historical data that closely match the current context $x$. We then filter the futures of these past conversations based on which have the highest percentage of control words.
\begin{align*}
\centering
\mathtt{KNN_x}&=\mathtt{faiss}(M(x), M(p^{(i,j)}), k)\\
f^*&=\mathtt{argmax}([\mathtt{count}(f^{(i,j)}, \mathcal{W})]_{f^{(i,j)} \in \mathtt{KNN_x}})\\
\tilde{s}_{t+1}&=f^*[0], f^*=\{s_1, u_1, ..., s_T, u_T\}
\end{align*}
In the above equations, the $\mathtt{count}$ function counts the number of control words $\mathcal{W}$ in the future $f^{(i,j)}$. The reference response $\tilde{s}_{t+1}$ is simply the first utterance of the retrieved future.

\subsection{Guided Futures of the Past (FOP-guided)}

Now that we have a candidate reference response $\tilde{s}_{t+1}$, we can guide the language model towards generating a similar response. To do this, we modify the logits from the language model to encourage generation of the control words or similar words. We start with the first word $w_0$ in $\tilde{s}_{t+1}$ and upweight logits in a way similar to DBS \cite{pascual2020directed} using similarity of GloVe vector embeddings:
\begin{align*}
\centering
l'_i&=l_i + \lambda \cdot \mathtt{min}(0, \mathtt{cos}(\gamma(t_i), \gamma(w_j)))^2, 
\end{align*}
where $\gamma$ represents GloVe embeddings, $t_i$ is the $i$th token of the language model's vocabulary $V$, $w_j$ is the current reference word, and $\lambda$ is a hyperparameter specifying how much weight to put on generating words similar to $w_j$.

With this approach, we observed that sometimes the model got stuck on the first word and never moved on to later words. To enable more flexible control, instead of requiring every word to be generated before moving on to the next word, we include a window of size $q$ and increase the logits of each word in the window, with a decay multiplier of $\frac{1}{2^i}$, $i \in q$. If any of the words in the window have been generated, the window is shifted beginning from the generated word with the same window size of $q$. The process repeats until the full response has been generated.

The decay multiplier is used to encourage the model to generate earlier words in the reference response and not skip words unless it’s highly likely. We generate $N$ such responses using this method and include an additional ranking step to select the best one. We first sort by the number of control words in the generated response. If multiple responses generate the highest number of control words, we sort by the loss from the model and select the response with the lowest loss $l$:
\begin{align*}
\centering
\tilde{\mathcal{R}}_{t+1}&=\{M^j(l') | j \in [1,...,N]\}\\
s^*&=\mathtt{max}([\mathtt{count}(r, \mathcal{W})]_{r \in \tilde{\mathcal{R}}_{t+1}})\\
\hat{\mathcal{R}}_{t+1}&= \{r | \mathtt{count}(r, \mathcal{W}) = s^*, r \in \tilde{\mathcal{R}}_{t+1}\}\\
r_{t+1}&=\mathtt{argmin}([\mathtt{loss}(r)]_{r \in \hat{\mathcal{R}}_{t+1}}),
\end{align*}
where $\tilde{\mathcal{R}}_{t+1}$ is the set of $N$ generated responses, using a model with logits $l'$. The final generated response $r_{t+1}$ is selected based on the two-step ranking process. None of the other approaches include this ranking component.

\section{Experimental setup}
\subsection{Task-Oriented Dialogue Datasets}
Our problem and approach are applicable to any general dialogue control setting. In our experiments, we controlled the customer in task-oriented dialogue. This is useful for constructing a customer bot that imitates real-life customers. By controlling the customer simulator (for example through control words), we can develop a training environment for coaching customer service agents in a variety of diverse situations. For all datasets, we select control words from the utterances of the customer by selecting the top $M$ ranked words based on tf-idf. For some real-world applications, control words can also be manually selected by a designer.
\paragraph{MultiWoz 2.3:} The first dataset we evaluate on is MultiWoz 2.3 \cite{han2020multiwoz}, which is widely used in the dialogue community. The dataset has over 10K dialogues and 5 domains.
\paragraph{TaskMaster-3:} The second is another commonly used task-oriented dialogue dataset TaskMaster-3 \cite{byrne-etal-2019-taskmaster}. This dataset has 23,757 dialogues in the movie ticketing domain.
\paragraph{Action-Based Conversations Dataset (ABCD):} The final dataset \cite{chen-etal-2021-action} includes a set of agent-customer conversations focused on solving customer problems. The dataset contains over 10k dialogues and is also focused on one domain.

\subsection{Baselines}
\paragraph{$\mathbfcal{W}_{\text{first}}$:} The first baseline is a naive approach that outputs all control words in the first response of the conversation and nothing afterwards, which means words are not appropriately timed.
\paragraph{Fine-tuned:} This approach simply generates responses using the fine-tuned language model $M$.
\paragraph{Prompt:} This method is based on prompting approaches \cite{li2021prefix,ribeiro2018semantically,jiang2020can,madotto2021few}. Because we focus on the plug-and-play setting, we simply append control words to the beginning of the context and generate using this modified input. 
\paragraph{Directed Beam Search (DBS):} This is a constrained text generation approach \cite{pascual2020directed}, in which keywords are generated using logit modification and beam search. It is not optimized for long-term control and is highly dependent on the ordering of control words.
\paragraph{Constrained Sentence Generation by Metropolis-Hastings Sampling (CGMH):} This method \cite{miao2019cgmh} is based on stochastic search methods that insert, delete, and replace words in a sentence with the requirement that control words need to be present. It is neither optimized for long-term generation of control words nor forward generation and is particularly susceptible to aggressively generating all control words in a single response. It was also originally applied to the task of keyword-to-phrase generation so we adapted it to dialogue generation by prompting the language model with the dialogue context and also replaced a bidirectional RNN model with our transformer-based model.

\section{Results}

\subsection{Aggregated Results}
\label{sec:summary_stats}
\begin{table}[t]
\small
\centering
\begin{tabular}{c|ccc|c} \toprule 
    \textbf{Methods} & \textbf{LT-} & $\bm{f}$\textbf{1-} & \textbf{Human} & \textbf{Overall}\\ 
    & \textbf{SR} & \textbf{score} & \textbf{eval} & \textbf{average}\\ \midrule
Prompt & 0.23 & 0.34 & \textbf{0.87} & 0.48\\
DBS & 0.42 & 0.28 & 0.72 & 0.47\\
CGMH & \textbf{0.90} & 0.17 & 0.3 & 0.46\\ \midrule
FOP-retrieval & 0.82 & 0.39 & 0.82 & \textbf{0.68}\\
FOP-guided & 0.74 & \textbf{0.41} & 0.81 & 0.67\\ \midrule
\end{tabular}
\caption{Summary table of results, including long-term success rate (LT-SR) from Figure \ref{fig:abcd_success_rate} averaged over datasets for 9 control words, F1-score from the overall F1 column of Table \ref{table:f1_score} that averages F1 over datasets, and human eval from Table \ref{table:human_eval} averaged over all metrics and divided by 5 to get a number between 0 and 1.}
\label{tab:summary_stats}
\end{table}
We begin by presenting a top-level overview of our main baselines and methods because each evaluation metric captures a different aspect of performance.  Table~\ref{tab:summary_stats} includes averaged scores across tasks, parameters, and/or metrics for the main results in Tables \ref{table:f1_score} and \ref{table:human_eval} and Figure \ref{fig:abcd_success_rate}. These include results of our two proposed automatic metrics of long-term success rate and control word F1-score (Section ~\ref{sec:metrics}) as well as human-evaluated quality metrics (Section ~\ref{sec:human_eval}). In subsequent sections, we will examine each of these results more closely.

The key insight in these aggregated results is that while FOP-based methods are not always the best-performing system for each metric, they are consistently the most reliable. Specifically, CGMH has high success rate, but lowest F1 and human scores. Prompt, on the other hand has the highest human evaluation scores but the worst success rate. This is not too surprising. It is, after all, an unmodified language model, so it should be fluent and on topic when viewed by a human. However, given its extremely low success rate, it is not viable for long-form controlled generation. In contrast, FOP-based methods are either the top 1 or 2 performing system across all summary statistics.

\begin{table*}[t]
\small
\centering
\begin{tabular}{c|ccc|ccc|ccc|c} \toprule 
    \multicolumn{1}{c}{\textbf{Methods}} & \multicolumn{3}{c}{\textbf{MultiWoz 2.3}} & \multicolumn{3}{c}{\textbf{TaskMaster-3}} & \multicolumn{3}{c}{\textbf{ABCD}} & \multicolumn{1}{c}{\textbf{Overall}} \\
     & {$p$} & {$r$} &
    {$f1$} & {$p$} & {$r$} &
    {$f1$} & {$p$} & {$r$} &
    {$f1$} & $\mathtt{avg}(f1)$ \\ \midrule
    $\mathcal{W}_{first}$ & 0.25 & 0.18 & 0.21 & 0.22 & 0.19 & 0.2 & 0.29 & 0.24 & 0.27 & 0.23\\
    Fine-tuned & \textbf{0.64} & \textbf{0.23} & \textbf{0.34} & \textbf{0.82} & 0.34 & 0.48 & 0.68 & 0.13 & 0.22 & 0.35\\ \midrule
    Prompt & 0.45 & 0.18 & 0.25 & 0.81 & 0.36 & 0.49 & \textbf{0.69} & 0.18 & 0.29 & 0.34\\
    DBS & 0.4 & 0.2 & 0.27 & 0.43 & 0.27 & 0.33 & 0.39 & 0.17 & 0.24 & 0.28\\
    CGMH & 0.27 & 0.18 & 0.21 & 0.17 & 0.03 & 0.05 & 0.27 & 0.22 & 0.24 & 0.17\\ \midrule
    FOP-retrieval & 0.38 & 0.18 & 0.25 & 0.68 & 0.38 & 0.49 & 0.65 & 0.33 & 0.44 & 0.39\\
    FOP-guided & 0.36 & 0.18 & 0.24 & 0.62 & \textbf{0.48} & \textbf{0.54} & 0.6 & \textbf{0.36} & \textbf{0.45} & \textbf{0.41}

\\ \bottomrule
\end{tabular}
\caption{Precision, recall, and F1-score for all methods on Multiwoz, TaskMaster, and ABCD. These metrics capture whether the approaches generate control words at the right time by using the control words in the ground truth response as a proxy. The last column is the macro $f1$-score average across all datasets.}
\label{table:f1_score}
\end{table*}

\subsection{Long-term Success Rate}
The first analysis involves comparing all methods on long-term success rate, which measures the percentage of control words in generated simulated roll-outs. To do this, we train a separate user model with the training dataset. We perform a roll-out per test example with 10 generated system responses and 10 generated user responses and compute the percentage of control words in the generated system responses. When counting the number of generated words, we compare word stems. 

\begin{figure}[t]
    \centering
    \includegraphics[width=0.47\textwidth]{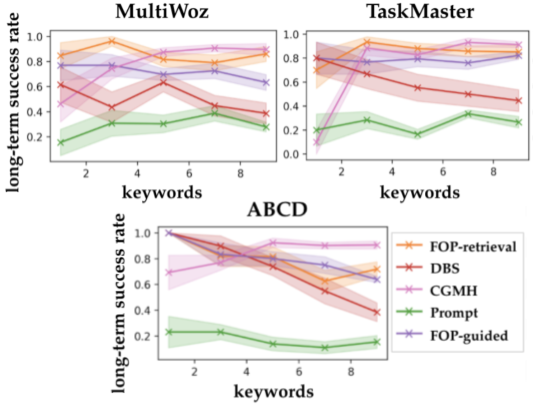}
    \caption{Long-term success rate computed on simulated roll-outs for MultiWoz, TaskMaster, and ABCD. Details on hyperparameters are in Appendix \ref{app:exp_details}.} \label{fig:abcd_success_rate}
\end{figure}

Figure \ref{fig:abcd_success_rate} shows the performance of all approaches when varying the number of control words. Both of our approach variants (FOP-retrieval and FOP-guided) have higher success rates than Prompt and DBS. Prompt is the method with the lowest performance because including the control words at the beginning without any re-training doesn't provide the model with sufficient information to generate the control words. DBS does well when there is only a few control words but struggles as the number of control words increases. This is because DBS is not able to filter out words that are irrelevant at the current time step and instead simply tries to generate the words one by one. This method is also unable to handle words when not in the exact order it should appear. 

FOP-retrieval, in some cases, has higher performance than FOP-guided because it will get all keywords in the retrieved response correct. FOP-guided can choose to ignore these keywords if the LM overrides it. So, we would expect FOP-retrieval to do better on this metric, compared to FOP-guided. We also include an ablation experiment in Appendix~\ref{app:ablation} to analyze the effect of removing the sliding window in FOP-guided. 
CGMH seems to do well on long-term success rate, but human evaluation (Section \ref{sec:human_eval}) results reveal that the generated responses are not very fluent. This method is one that can game previous evaluation metrics, as it tends to condense many or all control words into one utterance. Thus, these approaches are better evaluated through the next set of metrics: precision, recall, and F1-score.

\begin{table}[t]
\centering
\small
\begin{tabular}{r|lllll} \toprule
    {\textbf{Methods}} & {\textbf{FL}} & {\textbf{CC}} &
    {\textbf{RL}} &
    {\textbf{CO}} & {\textbf{DV}}\\ \midrule
    DBS & 4.60\textsuperscript{*} &                      3.65\textsuperscript{$\dagger$} &            \textbf{3.80} &       2.90 &       3.10\textsuperscript{$\dagger$} \\
    CGMH & 1.70\textsuperscript{$\dagger$} &                      1.24\textsuperscript{$\dagger$} &            1.52\textsuperscript{$\dagger$} &       1.12\textsuperscript{$\dagger$} &       1.82\textsuperscript{$\dagger$} \\ \midrule
    FOP-retrieval & \textbf{4.81} &                      \textbf{4.77} &            3.63 &       2.82 &       4.35 \\
    FOP-guided & 4.36\textsuperscript{$\dagger$} &                      4.53\textsuperscript{*} &            3.77 &       \textbf{3.12} &       \textbf{4.47} \\ \midrule[2pt]
    \rowcolor{Gray}
    Prompt & 4.87 &                      4.98 &            4.30 &       4.22 &       3.42 \\
    \rowcolor{Gray}
    True & 4.88 &                      4.90 &            4.83 &       4.92 &       4.80
\\ \bottomrule
\end{tabular}

\caption{Human evaluation of simulated roll-outs. \textbf{FL}: fluency; \textbf{CC}: control-consistency; \textbf{RL}: relevance; \textbf{CO}: Coherence; \textbf{DV}: diversity.  
\textsuperscript{*} and \textsuperscript{$\dagger$} indicate significant differences from the best result in that column (bolded, excluding True and Prompt) with p-value $<0.05$ and $<0.001$ respectively, using Welch's t-test.
Annotators rated fluency, control-consistency, and relevance per response, while coherence and diversity were annotated per conversation. All metrics are on a scale of 1 to 5.}
\label{table:human_eval}
\end{table}

\subsection{Control Word P/R/F1}
\label{sec:bleu_bert}

We now measure how well the approaches generate control words at the right time using precision, recall, and F1-score. Table \ref{table:f1_score} compares these metrics on all datasets. We see that, on average across all datasets, FOP-guided gets higher F1-scores compared with baseline methods. This is because by retrieving similar futures, we are able to guide the language model towards generating control words at appropriate points in the conversation. FOP-guided does worse on MultiWoz because the dataset contains more domains and has much more variety in the conversations. This diversity makes it hard for retrieval-based methods to successfully find similar conversations to guide generation.

The naive approach $\mathcal{W}_\text{first}$ gets low recall and precision since it only outputs the control words at the first utterance. Similar to $\mathcal{W}_\text{first}$, CGMH gets low F1-scores because it generates many control words early in the conversation rather than at a natural time. DBS also does not do well on these evaluation metrics as it is highly affected by the order of control words, while our method is able to retrieve similar futures to generate appropriate words at the current time step. Finally, Prompt does well on precision but not on recall as it's not explicitly guided to generate the control words.

\subsection{Human Evaluation}
\label{sec:human_eval}
Finally, we rate all methods on human evaluation. We follow recent work on good evaluation practices for text generation approaches \cite{karpinska2021perils}. Further details are in Appendix \ref{sec:human_evaluation_details}.

\noindent \textbf{Fluency}: Is the response fluent and grammatical?\\
\noindent \textbf{Control consistency}: \textit{When} control words appear in the response, are they appropriately used?\\
\noindent \textbf{Relevance}: Is the response a natural reply to the previous utterance in the conversation?\\
\noindent \textbf{Coherence}: Are all of the system responses in the conversation coherent with respect to each other?\\
\noindent \textbf{Diversity}: Is there diversity in the system responses of the conversation?

Two raters annotated each example, and agreement was measured using Krippendorff’s alpha for each of the 5 metrics (0.84, 0.74, 0.82, 0.76, 0.67). We present results in Table \ref{table:human_eval} for all five approaches as well as for the ground truth conversation. We focus on comparisons between DBS, CGMH, and the FOP methods, as these were the methods that performed comparably on control metrics (at least 40\% on long-term success rate) and thus are reasonable baselines for long-term control.

CGMH consistently gets low scores across all metrics. Compared to DBS, FOP-guided performs similarly on fluency, relevance, and coherence but much better on control-consistency and diversity, which could be because retrieval helps decide naturally what to say throughout the conversation. FOP-guided is at least as good as FOP-retrieval on relevance, coherence, and diversity, while only slightly worse on fluency and control-consistency. This is because FOP-guided uses the context and retrieved sentence to \textit{generate} a response, while FOP-retrieval selects an already fluent historical response. Overall, human evaluation results highlight that both of our proposed methods generate realistic, coherent text, while also generating a high percentage of control words.

\section{Conclusion}

In this paper, we propose the problem of constrained dialogue generation, which involves controlling dialogue responses such that a set of control words appear at some point in the future of the conversation. We propose a new set of metrics as well as a novel method that leverages retrieval of relevant conversations to control future generated responses. We show on three datasets that our method outperforms several constrained text generation baselines on quantitative metrics as well as human evaluation. As far as we are aware, this is the first work to address the problem of long-term control for dialogue generation.

\section{Acknowledgments}

We thank S.R.K Branavan and Derek Chen for their insightful feedback. We thank Tianyi Zhang for his starting code that we built upon in this work. We also want to thank Ethan Elenberg, Felix Wu, Clemens Rosenbaum, Sam Altschul, David Sontag, and the rest of the ASAPP research team for all of their feedback in making this work stronger.

\bibliography{anthology,custom}
\bibliographystyle{acl_natbib}

\clearpage
\appendix

\section{Appendix}
\label{sec:appendix}

\subsection{Additional results}

\subsubsection{Ablation of window in FOP-guided}
\label{app:ablation}
We ran ablation experiments comparing FOP-guided with a version without the sliding window. Table \ref{tab:ablation_table} includes the results for all of the baselines on the most difficult setting for ABCD (9 control words).

\begin{table}[h]
\small
\centering
\begin{tabular}{c|c} \toprule 
    \textbf{Methods} & \textbf{Long-term}\\ 
    & \textbf{success rate}\\ \midrule
Prompt & 0.15 \\
DBS & 0.38\\
CGMH & 0.91 \\ \midrule
FOP-retrieval & 0.72 \\
FOP-guided & 0.69 \\ 
FOP-guided (no-window) & 0.56 \\ \midrule
\end{tabular}
\caption{Ablation experiment for the most difficult setting in ABCD (9 control words). FOP-guided without a sliding window performs worse on long-term success rate.}
\label{tab:ablation_table}
\end{table}

\begin{figure}[h]
    \centering
    \includegraphics[width=0.4\textwidth]{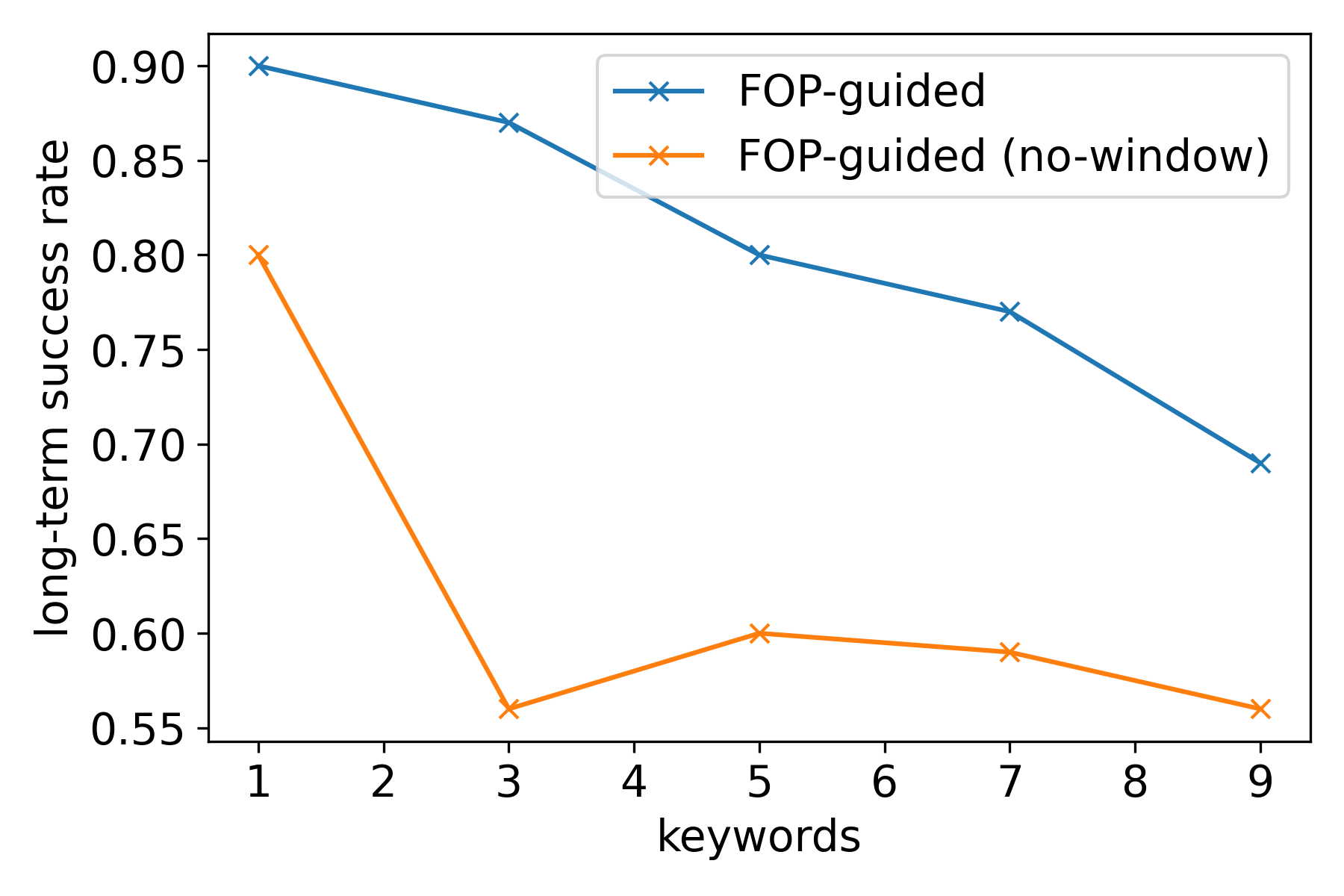}
    \caption{Long-term success rate on ABCD, comparing FOP-guided and FOP-guided without a sliding window.} \label{fig:ablation_graph}
\end{figure}

Our approach FOP-guided gets more than 10\% more control words in simulated rollouts, compared with FOP-guided without the window approach, which highlights the usefulness of the sliding window component. We also compare the two FOP-guided variants when varying the number of control words and see that FOP-guided consistently performs better (Figure \ref{fig:ablation_graph}).

\subsection{Example simulations on ABCD}

In Tables \ref{table:abcd_examples_fop_guided}, \ref{table:abcd_examples_fop_retrieval}, \ref{table:abcd_examples_dbs}, \ref{table:abcd_examples_cgmh}, and \ref{table:abcd_examples_prompt}, we show some example simulations on the ABCD dataset using a trained agent model for each of the methods.

\subsection{Experiment details}
\label{app:exp_details}
We did a hyperparameter search over the following lambda values $\{0,5,10,15,20,25\}$ for all datasets. On both ABCD and MultiWoz, the best hyperparameter for FOP-guided was $\lambda=15$ and for DBS, it was $\lambda=20$. For TaskMaster, the best hyperparameter for FOP-guided was $\lambda=10$ and for DBS, it was $\lambda=15$. CGMH was run with the recommended hyperparameters from the authors.

For all datasets, we used the number of candidate generations for FOP-guided as $N=10$ and the window size for logit modification as $q=4$. The number of examples used for multiple splits of each dataset is as follows: For the ABCD dataset, we used 8034 conversations for training and 1004 conversations each for dev and test splits. In the Multiwoz dataset, we used 8438, 1000, 1000 as train, dev and test splits respectively. Finally, for the Taskmaster-3 dataset, we used 16629, 3564, 3564 as train, dev and test datasets respectively.

We used the GPT2-medium model from the hugging-face repository as the pre-trained language model for all of our experiments. This model contains 345M parameters.

For all our experiments, we used a p3.2xlarge EC2 instance. This instance has one GPU with 16GB capacity and 61GB of RAM. Out of all of our experiments, simulated long-term success rate experiments took the most amount of GPU hours to run. Altogether it took somewhere between 24-36 GPU hours to complete all the experiments.

\subsection{Human evaluation setting details}
\label{sec:human_evaluation_details}
We recruited four trained annotators to evaluate generated conversations on the following five metrics, each on a scale of 1 to 5. We split up the examples across the four annotators such that each example was judged by two annotators. We included the ground truth conversation as an additional baseline to act as an upper bound. To ensure the ratings would be high-quality, we provided a rubric, included below, for each metric with examples for different ratings, did an initial pilot for a few sample conversations, and provided a reference sheet to help calibrate the ratings across annotators.

\subsubsection{Rubric}
Evaluate generated conversations on a few metrics, each on a scale of 1 to 5:\\

\noindent \textbf{[utterance-level] Fluency: Is this response fluent and grammatical?}

\begin{itemize}
\item 1: Generated responses do not make any sense, English-wise and grammar-wise, which could include misspelled words, no transition words, limited punctuation, skipped words, etc (e.g., “the figh help order”)

\item 3: Generated responses have some good English so you can make out what is being said but it’s not well-formed sentences (e.g., “will you help order”)

\item 5: Generated responses have perfect English and perfect grammar. Customers can use lower-case text as less-formal style so first-letter capitalization is not necessary (e.g., “can you help me refund my order?”)
\end{itemize}

\noindent \textbf{[utterance-level] Relevance: Is this response a natural reply to the previous utterance in the conversation?}
\begin{itemize}
\item 1: The generated response is not at all relevant to the conversation context/history (e.g., when asked for account id: “I can’t get my promo code”)

\item 3: The generated response is somewhat relevant to the conversation context/history but not the best fit (e.g., when asked for account id: “No”)

\item 5: The generated response is perfectly relevant and a great response to the conversation context/history (e.g., when asked for account id: “Account ID: 3425435”)
\end{itemize}
\noindent \textbf{[utterance-level] Control-consistency: If control words appear in this response, are they appropriately used?}
\begin{itemize}
\item 1: When used, the control words (which are uppercased) make no sense in the generated responses. They are fully forced into the responses (e.g., “TODAY account id: 435650”) 

\item 3: When used, the control words (which are uppercased) make some sense in the generated responses but are not super smooth (e.g., “I need help with my order, can you help TODAY?“)

\item 5: When used, the control words (which are uppercased) are perfectly and naturally used in the generated responses (e.g., “TODAY, I want to buy a shirt. Can you help me?”) 
\end{itemize}
\noindent \textbf{[conversation-level] Diversity: Is there diversity in the customer responses of the conversation?}
\begin{itemize}
\item 1: Almost all of the responses are repetitive and have no diversity (e.g., “ok” “ok” “ok” “thanks”)

\item 3: Some of the generated responses provide diversity while many do not (e.g., “I want to buy a shirt” “can you help me with this?” “ok” “thanks”)

\item 5: All of the generated responses are diverse and provide a variety of interesting words through the conversation. The customer can still say ok and thanks but it shouldn’t happen all the time and has to be appropriate for that point in the conversation (e.g., “I want to buy a shirt” “can you help me with this?” “ok sure” “thank you very much for your help!”)
\end{itemize}
\noindent \textbf{[conversation-level] Coherence: Are all of the customer responses in the conversation coherent with respect to each other?}
\begin{itemize}
\item 1: All of the responses have very little relation when seen together (e.g., “I want to buy a shirt”, “can you help me with my promo code”, “my subscription expired”)

\item 3: Some of the generated responses in a conversation make sense together and some don’t (e.g., “I want to buy a shirt” “I have a promo code for buying this shirt” “my subscription expired”)

\item 5: All generated responses make complete sense together (e.g., “I want to buy a shirt”, “I have a promo code for buying this shirt”, “thank you for helping me with this purchase!”)
\end{itemize}

\begin{table*}[t]
\centering
\rowcolors{2}{gray!25}{white}
\begin{tabular}{p{0.15\linewidth}p{0.75\linewidth}} \toprule
    \rowcolor{white} \multicolumn{2}{c}{FOP-guided} \\ \toprule
    \rowcolor{white} \multicolumn{2}{c}{\textbf{Control words to generate:} valid credit make minute try card purchase trying worked} \\ \toprule
    \rowcolor{white} \textbf{Speaker} & \textbf{Utterance} \\ \midrule
    AGENT &  Hello, thank you for contacting AcmeCorp, how may I help you today? \\
 CUSTOMER & Hi I was \textbf{trying} to \textbf{make} a \textbf{purchase} but for some reason my \textbf{credit} \textbf{card} keeps getting rejected \\
 AGENT &  I would happy to help. \\
 CUSTOMER & I got an email from you guys about a great sale and it listed this code and it said it was \textbf{invalid} \\
 AGENT &  Lets try a few things. First can you check the expiration date on your card. \\
 CUSTOMER & No its \textbf{valid} thats why I am confused. \\
 AGENT &  Can you try to re-enter the card information one more time? \\
 CUSTOMER & Ok just a \textbf{minute} please \\
 AGENT &  Alright, let me know if that helps. \\
 CUSTOMER & Ok I did its s ordering right now. That \textbf{worked} and it says it is correct. \\
 AGENT &  Great, is there anything else I can help you with today?  Is there anything else? \\
 CUSTOMER & That is all I needed thanks \\
 AGENT &  Your welcome. Thank you for contacting us. Have a great day. \\
 CUSTOMER & you too \\ \bottomrule
\end{tabular}
\caption{Example simulation on the ABCD dataset using the FOP-guided method}
\label{table:abcd_examples_fop_guided}
\end{table*}

\begin{table*}[t]
\centering
\rowcolors{2}{gray!25}{white}
\begin{tabular}{p{0.15\linewidth}p{0.75\linewidth}} \toprule
    \rowcolor{white} \multicolumn{2}{c}{FOP-retrieval} \\ \toprule
    \rowcolor{white} \multicolumn{2}{c}{\textbf{Control words to generate:} valid credit make minute try card purchase trying worked} \\ \toprule
    \rowcolor{white} \textbf{Speaker} & \textbf{Utterance} \\ \midrule
AGENT &  Hello, thank you for contacting AcmeCorp, how may I help you today? \\
 CUSTOMER & Hi! I was \textbf{trying} to \textbf{make} a \textbf{purchase} but for some reason my \textbf{credit} \textbf{card} keeps getting rejected \\
 AGENT &  I would happy to help. Lets see if we can troubleshoot this issue. \\
 CUSTOMER & ok let me know what i should \textbf{try} \\
 AGENT &  Lets try to make sure your card is not expired. \\
 CUSTOMER & no. its \textbf{valid}. Thats why I am confused \\
 AGENT &  First can you try to log out and log back in. \\
 CUSTOMER & Ok let me give that a shot. \\
 AGENT &  Okay, next can you check the expiration date on your card. \\
 CUSTOMER & Ok hang on one \textbf{minute} let me \textbf{try} that \\
 AGENT &  No problem. \\
 CUSTOMER & Tried that. It didn't \textbf{work} either \\
 AGENT &  Can you try to check the expiration on your card? \\
 CUSTOMER & Let me \textbf{try} that \\
 AGENT &  Alright. \\
 CUSTOMER & Well that kinda helped. I do not know why \\
  \bottomrule
\end{tabular}
\caption{Example simulation on the ABCD dataset using the FOP-retrieval method}
\label{table:abcd_examples_fop_retrieval}
\end{table*}

\begin{table*}[t]
\centering
\rowcolors{2}{gray!25}{white}
\begin{tabular}{p{0.15\linewidth}p{0.75\linewidth}} \toprule
    \rowcolor{white} \multicolumn{2}{c}{DBS} \\ \toprule
    \rowcolor{white} \multicolumn{2}{c}{\textbf{Control words to generate:} valid credit make minute try card purchase trying worked} \\ \toprule
    \rowcolor{white} \textbf{Speaker} & \textbf{Utterance} \\ \midrule
AGENT &  Hello, thank you for contacting AcmeCorp, how may I help you today? \\
 CUSTOMER & I would like to know if there is any reasonable way for \textbf{valid}ating an order? \\
 AGENT &  I would happy to look into this for you. \\
 CUSTOMER & I already got a \textbf{credit} for a return and would like to correct it. \\
 AGENT &  May I have your full name or account id please? \\
 CUSTOMER & Crystal Minh \\
 AGENT &  Thank you Crystal. \\
 CUSTOMER & Can you \textbf{make} sure that this is correct? \\
 AGENT &  One moment please. \\
 CUSTOMER & ok \\
 AGENT &  What is your username, email address, and Order ID? \\
 CUSTOMER & crystalm \textbf{minute} \\
 AGENT &  Thanks Crystal.  I need to verify your purchase. \\
 CUSTOMER & crystalm \textbf{minute}@email.com \\
 AGENT &  Your refund is in progress and you initiated it online. \\
 CUSTOMER & Okay. I will \textbf{try} it out now. \\
  \bottomrule
\end{tabular}
\caption{Example simulation on the ABCD dataset using the DBS method}
\label{table:abcd_examples_dbs}
\end{table*}

\begin{table*}[t]
\centering
\rowcolors{2}{gray!25}{white}
\begin{tabular}{p{0.15\linewidth}p{0.75\linewidth}} \toprule
    \rowcolor{white} \multicolumn{2}{c}{CGMH} \\ \toprule
    \rowcolor{white} \multicolumn{2}{c}{\textbf{Control words to generate:} valid credit make minute try card purchase trying worked} \\ \toprule
    \rowcolor{white} \textbf{Speaker} & \textbf{Utterance} \\ \midrule
AGENT &  Hello, thank you for contacting AcmeCorp, how may I help you today? \\
 CUSTOMER & My\textbf{valid} \textbf{credit} \textbf{card} \textbf{make} me \textbf{purchase} \textbf{minute} we \textbf{try} to \textbf{credit} \textbf{card} to \textbf{purchase}. Have been \textbf{trying} this \textbf{worked} \\
 AGENT &  I would happy to look into this for you. \\
 CUSTOMER & I was wondering about this method of\textbf{valid} \\
 AGENT &  May I have your full name or account id please? \\
 CUSTOMER & sure its a\textbf{valid} \\
 AGENT &  Thank you may I please have your full address? \\
 CUSTOMER & 9 Guess\textbf{valid} \\
 AGENT &  What method of payment would you like to use? \\
  \bottomrule
\end{tabular}
\caption{Example simulation on the ABCD dataset using the CGMH method}
\label{table:abcd_examples_cgmh}
\end{table*}

\begin{table*}[t]
\centering
\rowcolors{2}{gray!25}{white}
\begin{tabular}{p{0.15\linewidth}p{0.75\linewidth}} \toprule
    \rowcolor{white} \multicolumn{2}{c}{Prompt} \\ \toprule
    \rowcolor{white} \multicolumn{2}{c}{\textbf{Control words to generate:} valid credit make minute try card purchase trying worked} \\ \toprule
    \rowcolor{white} \textbf{Speaker} & \textbf{Utterance} \\ \midrule
AGENT &  Hello, thank you for contacting AcmeCorp, how may I help you today? \\
 CUSTOMER & Hi I’m having a problem with my account \\
 AGENT &  I would happy to help. \\
 CUSTOMER & The \textbf{credit} \textbf{card} keeps getting rejected \\
 AGENT &  Lets see if we can try a few things. \\
 CUSTOMER & ok \\
 AGENT &  Can you try entering the information again? \\
 CUSTOMER & Okay I did but it still doesn’t \textbf{work} \\
 AGENT &  Next can you try logging out and back in? \\
 CUSTOMER & Ok hold on \\
 AGENT &  Let me know if that helps. \\
 CUSTOMER & hey that \textbf{worked}! \\
  \bottomrule
\end{tabular}
\caption{Example simulation on the ABCD dataset using the Prompt method}
\label{table:abcd_examples_prompt}
\end{table*}

\end{document}